# Design and Development of Rule-based open-domain Question-Answering System on SQuAD v2.0 Dataset

Pragya Katyayan[1], Nisheeth Joshi[2]

{pragya.katyayan@outlook.com, nisheeth.joshi@rediffmail.com}

[1,2]Department of Computer Science, Banasthali Vidyapith, Rajasthan.

**Abstract.** Human mind is the palace of curious questions that seek answers. Computational resolution of this challenge is possible through Natural Language Processing techniques. Statistical techniques like machine learning and deep learning require a lot of data to train and despite that they fail to tap into the nuances of language. Such systems usually perform best on close-domain datasets. We have proposed development of a rule-based open-domain question-answering system which is capable of answering questions of any domain from a corresponding context passage. We have used 1000 questions from SQuAD 2.0 dataset for testing the developed system and it gives satisfactory results. In this paper, we have described the structure of the developed system and have analyzed the performance.

## 1 Introduction

Curiosity is the fuel for discovery, inquiry and learning. Human's never-ending thirst for knowledge has always existed. However, being an expert of all trades is a next to impossible task for any person. In such case, machines seem to be a feasible option to fulfil the responsibility of being an expert of any (or all) field(s). using Natural Language Processing (NLP), we can enable computers to understand natural language questions and fetch an answer for it. The phenomenon is called Question-Answering (QA). QA is an AI-complete problem which has been one of the toughest challenges of NLP and AI. There have been researches in this field which have tried to develop a decent QA system with different techniques including rule-based [Riloff and Thelen (2000), Jain and Dodiya (2014), Veisi and Shandi (2020)] and statistical techniques [Hermajakob (2001), Ko et al. (2007), Iyyer et al. (2014)]. However, they all have their own drawbacks, for instance- most of the developed systems are not open-domain, so, they perform well only for the domains that they have been trained on. The state-of-the-art system was developed by IBM in the name of IBM Watson which was developed to answer natural language questions in the game show Jeopardy and beat human experts. It fulfilled its goal and later became an expert system. Nevertheless, research in this direction has continued. Statistical techniques like Machine Learning (ML) and Deep Learning (DL) have their own disadvantages, one of which is the requirement of huge dataset for training purposes. Despite training on large datasets, these corpus-based techniques fail to understand the nuances of language most of the time. Hence, we have tried to take a rule-based approach to solve this problem. We propose a rule-based open-domain question-answering system capable of answering questions based on rules. The proposed system would consist of three modules for question processing, sentence extraction and answer processing. The modules would work in a pipeline to accomplish the task and will provide answers for questions. This paper provides detailed working of all the modules in the proposed system.

The paper has been organized in the following manner: section 2 throws some light on the previous works done in this area, section 3 elaborates on all the modules and sub-modules of the developed system, section 4 discusses the experimental setup, section 5 discusses the results and gives an error analysis for few questions and system's predictions over two sample passages and section 6 concludes the work with some highlights for future works.

## 2 Literature Review

Talking about QA systems, the very first system that revolutionized the researchers in this field was IBM's Watson. Ferrucci et al. (2010) devised a computer system that completed against human geniuses at the game show Jeopardy! And defeated them. The system was capable of making strategic decisions while finding answers of quiz questions. It used Prolog for knowledge representation (Lally and Fodor, 2011) and also used evidence scoring algorithm by Murdock (2011). The answer merging and ranking framework was explained by Gondek et al. (2012). For Watson to be a successful open-domain QA system, it was crucial that it used a huge knowledge source (Fan et al., 2012). Kalyanpur et al. (2012b) argued that decomposable complex factoid questions are best to be broken down to simple forms to support QA systems. While answering a natural language question has its own complexities (Chu-Caroll et al., 2012), Watson was given the speed and accuracy to compete against human brains (Epstein et al., 2012).

Wang et al. (2014) have focused on developing an easily scalable portable system from Freebase while using Web Questions dataset having 5810 QA pairs. Embedding based rule extraction approach demonstrated by Yang et al. (2014) outperformed confidence-based approach. Semantic parsing framework for QA was proposed by Yih et al. (2015) which was capable of utilizing knowledge-base in a better manner. Seo et al. (2016) proposed the BiDAF network that has modules working in hierarchical fashion and is capable of obtaining a question-aware context display without summarizing. An end-to-end neural model was proposed by Wang & Jiang (2016) based on LSTM for Machine Comprehension. They used SQuAD dataset made by Rajpurkar et al. (2016) that has over 1 lakh questions asked by users over a group of 536 Wikipedia articles. There have been attempts of using RNN-based heuristics to make QA simpler than other attempts by Weissenborn et al. (2017). Reinforcement learning has also been sed to develop NN-based ReasoNets for machine comprehension by Shen et al. (2017). Tan et al. (2017) have worked on answer extraction model from docs and synthesizing them in form of an answer transfer-based learning for machine comprehension has been developed by Golub et al. (2017) on NewsQA dataset. SQuAD 2.0 was released with ~53000 fresh questions by Rajpurkar et al. (2018).

Riloff and Thelen (2000) developed Quarc, which was a rule-based system that reads small stories and finds sentences that presents themselves as best possible answer for that question. Textract was developed by Srihari & Li (2000) that extracts information for natural language QA. Hermajakob (2001) performed question classification and parsing using ML for QA. Ko et al. (2007) proposed probability-based graphical model for answer ranking in QA. Iyyer et al. (2014) proposed RNN model named QANTA for factoid QA on paragraphs. It could deal with cases where question words are vague and can't help much with answer extraction. Jain and Dodiya (2014) proposed rule-based architecture for QA system with a goal of extracting answers of medical questions rather than full documents or passages. A general why-how question answering system was developed by Pechshri and Piriyakul (2016) for answering questions of farmers. They used ML to determine type of questions. Jayalakshmi and Sheshasayee (2017) evaluated similarity scores of question-words and meaning between them to find correct answers. Cuteri et al. (2019) presented a closed-domain QA system for cultural heritage. Veisi and Shandi (2020) have designed and developed a medical QA system in Persian language using rule-based technique. Jiang et al. (2021) have used crawlers to use medical websites as data sources and used rule-based matching to classify and query questions on a medical QA system based on knowledge graph.

## 3 Structure of Rule-based Question-Answering model:

Our aim was to develop an open-domain QA model for English. The model would be able to take questions and contexts from any domain and be able to extract an answer for the question. The three main steps in any QA model are: query processing, sentence extraction and answer processing. The developed system also works in a similar manner. The working of each module is explained in the following sub-sections and Figure 1 describes each module.

### 3.1 Question processing module:

The question was thoroughly processed for all possible information that it could provide. The module preprocessed the questions to tokenize, normalize, lemmatize and removed stop words. Then, it extracted features (as mentioned in table 1) from the question that were capable of estimating candidate answers. We used Stanford CoreNLP as XML-RPC server to extract these features and rest of the programming was performed using Python 3.8.

**Table 1.** Features extracted from Questions

| Sl. No. | Question Features |
|---|---|
| 1 | Wh – words |
| 2 | Keywords |

| Sl. No. | Question Features |
|---|---|
| 3 | PoS |
| 4 | NER |

We are aware of the fact that questions can be rephrased to ask the same information. Hence, our next step was to perform question paraphrasing. It produced 5-10 different variations of the questions from which we again extracted keywords and added to our original keywords list. Apart from this, we observed that context passage may either contain exact words from the question or there may be similar words (or synonym) representing them. So, apart from these four features, we also extracted synonyms of question keywords using WordNet and created a map of each keyword with their synonyms.

This module was also responsible to analyze the wh-words and keywords of a question to judge the nature (or domain) of candidate answer. It had rules that processed the question structure and estimated whether question was asking for a person's name or a city's name or may be asking for a date or time of the year when a particular event happened. This information helped us in extracting answers in the next module.

**3.2 Sentence extraction module:**

This module performed multiple tasks ranging from context preprocessing to extracting the final sentence having a candidate answer. So, we divided all the tasks to several sub-modules as follows:

*3.2.1 Context preprocessing module:* This module tokenized, normalized, lemmatized and extracted keywords from the context passages.

*3.2.2 Sentence Extraction:* Using the presence of question keywords and their synonyms, this module used different extraction functions to extract most relevant sentence(s) from the context passage. Their relevance was based on the presence of a candidate answer in the sentence. To match the keywords in passage sentences, we first used exact matches of question keywords in the passage. Second, we matched synonyms of question keywords with context sentences. Thirdly, we used glove pretrained model of 300 dimensions to find cosine similarity between question keywords and context keywords. Most relevant sentences from all three steps were gathered and sent to the next sub-module.

*3.2.3 Feature extraction:* This module extracted PoS and NER values from the extracted sentence(s) and identifies important pieces of information from the sentence. We used StanfordCoreNLP to extract PoS and NER values for these sentences.

*3.2.4 Extraction and Scoring of Candiate Answers:* This module analyzed the domain of answer asked by the question (extracted in question processing module) and used different set of custom rules to look for relevant answers within the extracted sentences. For instance, if the question asked for a person's name as answer, then this module looked for names in the extracted sentences. It kept the sentences that had candidate answers and dropped the rest. Also, it gathered how many candidate answers were present in all sentences and maintained all the indices of all candidates in a separate list.

It was observed that most of the sentences in the context passage were long. So, in the event of multiple candidate answers in a single sentence, our module's next step was to match the nearest keywords (in a 3 to 5-word window) of the candidate answers with question keywords. The matching was done on the basis of exact match (90% weight), synonym match (70% weight), cosine similarity score >= 60% (60% weight). We prioritized exact matches between question keywords and extracted sentences by giving them 90% weightage. However, synonym match and cosine similarity matches were given 70% and 60% weight, respectively in the final score. The reason of this variation of weights was the fact that in quite a few cases we observed that synonyms and similar words out-scored exact matches, which was not encouraged for our system's performance. So, although their presence is valued, but on a slightly less magnitude than exact matches. These scores were added up and averaged to get a single score for every answer. The candidate answers with their match scores were stored in a separate list.

### 3.3 Answer processing module:

This module ranked the candidate answers based on their match scores with the question and produces the highest ranked answer in proper form as it appears in the context passage.

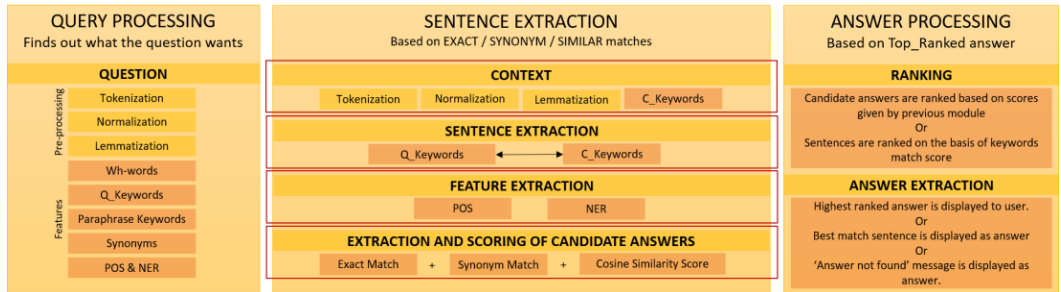

**Figure 1.** Three modules and sub-modules of the developed system.

### 3.4 Exception Handling:

There were some exceptional cases, where the rules in the scoring module were not able to pin point a particular answer in the sentence with highest matching score. In that case, the answer processing module presented with the sentence with highest matching score with the question as the answer. We assumed the sentence might have the correct answer, but due the complex and unpredictable nature of language (and in some cases the cause being <100% accuracy of PoS and NERs), our rules weren't able to catch the exact answer. Hence, if the user is presented with the highest match sentence, it would serve as a viable solution.

Also, there were few cases where question keywords didn't match with any sentence of the context passage by any criteria. For such cases, there were no exact match words or synonym matches and cosine similarity scores were less than the threshold of 60%. So, the rules we developed couldn't find any matching statements for that particular question. In such few cases, our system presented with a 'answer not found' message and requested the user to rephrase the question.

## 4 Experimental Setup

We developed the rules for the open-domain QA system and used SQuAD 2.0 dataset to test its performance. The SQuAD 2.0 dataset was developed by Rajpurkar et al. (2018) and it contains over 1 lakh questions from different domains which were collected from over 500 Wikipedia articles. It is a gold-standard dataset that has questions, their correct answers and the corresponding context passages as well.

We tested the developed QA system initially on 30 questions from 3 different domains of SQuAD 2.0 dataset. The system was continuously tweaked and tested until it gave 83% accuracy (answered 25/30 questions correctly, 4/30 got wrong answers while 1 question received a 'answer not found' message). Our next step was to test the system on a bigger set of questions. We took 1000 questions from SQuAD 2.0 dataset and tested the system's performance. The results are discussed in the next section.

## 5 Results and Discussion

We tested the developed rule-based system twice. First while developing the system, we tested it continuously on a set of 30 questions from different domains available in SQuAD 2.0 dataset. Our aim was to form rules that could cater all types of questions from different domains. After getting satisfactory results on a small dataset of 30 questions, we tested the system on a large dataset of 1000 questions from various domains. The system performed reasonably well and the results are presented in table 2. Table 3 gives the precision, recall, f-measure for both the test sets.

**Table 2.** The results of developed system on both test sets.

|  | Test Dataset size - 30 | | Test Dataset size - 1000 | |
| --- | --- | --- | --- | --- |
|  | **Total** | **Accuracy** | **Total** | **Accuracy** |
| **Correct answers** | 25 | 83.33% | 602 | **60.2%** |
| **Incorrect answers** | 4 | 13.33% | 390 | 39% |
| **No Answer** | 1 | 3.33% | 8 | 0.8% |

**Table 3.** The accuracy, precision, recall and F1-score of the system based on its performance.

|                          | Accuracy | Precision | Recall | F1-score |
|--------------------------|----------|-----------|--------|----------|
| **Test Dataset size - 30**   | 83.33    | 86.20     | 96.15  | 90.72    |
| **Test Dataset size - 1000** | **60.20**| 60.69     | 60.2   | 60.44    |

The developed system shows reasonable results with large dataset. Some of the answers predicted by the system are presented in table 4. We will analyze the predictions on 7 questions from two context passages to understand where the system fails and where it performs extraordinarily well.

**Table 4.** Examples of context, questions and answers as available in SQuAD 2.0 dataset and predicted answers by the developed RBQA system.

| Title | Content |
|---|---|
| **Context-1** | "Beyonce Giselle Knowles-Carter (born September 4, 1981) is an American singer, songwriter, record producer and actress. *Born and raised in Houston, Texas, she performed in various singing and dancing competitions as a child, and rose to fame <u>in the late 1990s</u>* as lead singer of R&B girl-group Destiny's Child. Managed by her father, <u>*Mathew Knowles*</u>, the group became one of the world's best-selling girl groups of all time. *Their hiatus saw the release of Beyonce's debut album, <u>Dangerously in Love</u> (2003), which established her as a solo artist worldwide, earned five Grammy Awards and featured the Billboard Hot 100 number-one singles 'Crazy in Love' and 'Baby Boy'."* |
| **Question-1** | **Who managed the Destiny's Child group?** |
| **Correct Answer** | Mathew Knowles |
| **Predicted Answer** | Mathew Knowles ✓ |
| **Question-2** | **What was the first album Beyonce released as a solo artist?** |
| **Correct Answer** | Dangerously in Love |
| **Predicted Answer** | "Their hiatus saw the release of Beyonce's debut album, <u>Dangerously in Love</u> (2003), which established her as a solo artist worldwide, earned five Grammy Awards and featured the Billboard Hot 100 number-one singles 'Crazy in Love' and 'Baby Boy'." ✓ |
| **Question-3** | **When did Beyonce start becoming popular?** |
| **Correct Answer** | in the late 1990s |
| **Predicted Answer** | 2003 ✗ |
| **Question-4** | **When did Beyonce rise to fame?** |
| **Correct Answer** | late 1990s |
| **Predicted Answer** | late 1990s ✓ |
| **Context-2** | "Following the disbandment of Destiny's Child in June 2005, she released her second solo album, B'Day (2006), which contained hits 'Deja Vu', 'Irreplaceable', and 'Beautiful Liar'. *Beyonce also ventured into <u>acting</u>, with a Golden Globe-nominated performance in <u>Dreamgirls</u> (2006), and starring roles in The Pink Panther (2006) and Obsessed (2009).* Her marriage to <u>*rapper Jay Z*</u> and portrayal of Etta James in Cadillac Records (2008) influenced her third album, I Am... Sasha Fierce (2008), which saw the birth of her alter-ego Sasha Fierce and earned a record-setting six Grammy Awards in 2010, including Song of the Year for 'Single |

| Title | Content |
|---|---|
| | Ladies (Put a Ring on It)'. Beyonce took a hiatus from music in 2010 and took over management of her career; her fourth album 4 (2011) was subsequently mellower in tone, exploring 1970s funk, 1980s pop, and 1990s soul. Her critically acclaimed fifth studio album, Beyonce (2013), was distinguished from previous releases by its experimental production and exploration of darker themes." |
| **Question-1** | **Which artist did Beyonce marry?** |
| **Correct Answer** | Jay Z |
| **Predicted Answer** | rapper Jay Z ✓ |
| **Question-2** | **For what movie did Beyonce receive her first Golden Globe nomination?** |
| **Correct Answer** | Dreamgirls |
| **Predicted Answer** | "Beyonce also ventured into acting, with a Golden Globe-nominated performance in Dreamgirls (2006), and starring roles in The Pink Panther (2006) and Obsessed (2009)." ✓ |
| **Question-3** | **After her second solo album, what other entertainment venture did Beyonce explore?** |
| **Correct Answer** | acting |
| **Predicted Answer** | "Following the disbandment of Destiny's Child in June 2005, she released her second solo album, B'Day (2006), which contained hits 'Deja Vu', 'Irreplaceable', and 'Beautiful Liar'." ✗ |

As we observe the first context, the very first question was asking for a person's name and got answered correctly. If we observe the context-1 passage we see that the name of the group is actually in the previous sentence of the one containing the correct answer. Since we prioritize exact matches, the most relevant sentence appears to the one having most keyword matches and, in this case, the correct answer – Mathew Knowles. In the next question, we see the system was not able to collect the exact answer, hence it presented the sentences with highest match score. If we observe closely, all the question keywords are present in the predicted sentence with one exception. The word 'first' is not available in the sentence, but the system is able to match it with a similar meaning word available in the sentence, i.e., debut. The next two questions are semantically similar but have different lexical representations. Interestingly, the predictions of both questions should be the same, but turned out different. 'Start becoming popular' and 'rose to fame' are semantically similar, but the context passage has 'start becoming popular', so the question with exact matches gets the right answer. However, the similar phrase fails to catch the correct sentence.

In the second context passage, the first question asks for a person's name and gets the correct answer. For the second question the system provides the sentence with most match-score and it has the correct answer. However, in the third question, all the keywords can be found in the predicted sentence, but it does not contain the answer. If we observe the context passage-2, the first sentence has all the keywords matching with the question keywords, but the second sentence has the correct answer. In such few cases, our hypothesis- of finding the keywords and the correct answer in the same sentence, struggles.

This analysis shows that our developed rule-based QA system performs reasonably well on open-domain questions and our hypothesis was correct most of the time. Considering the huge difference in resources, our system did well on the time complexity criteria.

## 6 Conclusion and Future Work

We developed a rule-based question-answering system with three main modules- question processing module, sentence extraction module and the answer processing module. The developed system presented satisfactory results on random open-domain questions from SQuAD 2.0 dataset. However, the result was not extraordinary. This gives us a glimpse of language complexity. Natural language is complex in its own way and can't be explained with either statistical methods or rules alone. A mix of both worlds is needed to get best results. In our future works, we will work on incorporating machine learning and deep learning methods in this model and will make a hybrid model that gets the benefits of both rule-based as well as statistical methods.